\documentclass[10pt,twocolumn,letterpaper]{article}

\usepackage{iccv}
\usepackage{times}
\usepackage{epsfig}
\usepackage{graphicx}
\usepackage{amsmath}
\usepackage{amssymb}
\usepackage{multirow}
\usepackage{bbding}
\usepackage{verbatim}

\usepackage[pagebackref=true,breaklinks=true,letterpaper=true,colorlinks,bookmarks=false]{hyperref}

\iccvfinalcopy 

\def\iccvPaperID{3409} 

\ificcvfinal\fi
\begin{document}

\title{DenseBody: Directly Regressing Dense 3D Human Pose and Shape From a Single Color Image}

\author{Pengfei Yao\dag\\
Cloudwalk\\
{\tt\small yaopengfei@cloudwalk.cn}
\and
Zheng Fang\dag\\
Cloudwalk\\
{\tt\small fangzheng@cloudwalk.cn}
\and
Fan Wu\\
Cloudwalk\\
{\tt\small wufan@cloudwalk.cn}
\and
Yao Feng\\
Shanghai Jiao Tong University\\
{\tt\small fengyao@sjtu.edu.cn}
\and
Jiwei Li\\
Cloudwalk\\
{\tt\small lijiwei@cloudwalk.cn}
}


\maketitle

\begin{abstract}
   Recovering 3D human body shape and pose from 2D images is a challenging task due to high complexity and flexibility of human body, and relatively less 3D labeled data. Previous methods addressing these issues typically rely on predicting intermediate results such as body part segmentation, 2D/3D joints, silhouette mask to decompose the problem into multiple sub-tasks in order to utilize more 2D labels. Most previous works incorporated parametric body shape model in their methods and predict parameters in low-dimensional space to represent human body. In this paper, we propose to directly regress the 3D human mesh from a single color image using Convolutional Neural Network(CNN). We use an efficient representation of 3D human shape and pose which can be predicted through an encoder-decoder neural network. The proposed method achieves state-of-the-art performance on several 3D human body datasets including Human3.6M, SURREAL and UP-3D with even faster running speed.
\end{abstract}

\renewcommand{\thefootnote}{}
\footnotetext{\dag Both authors contribute equally to this work.}

\begin{figure*}
\centering
\includegraphics[width=17cm]{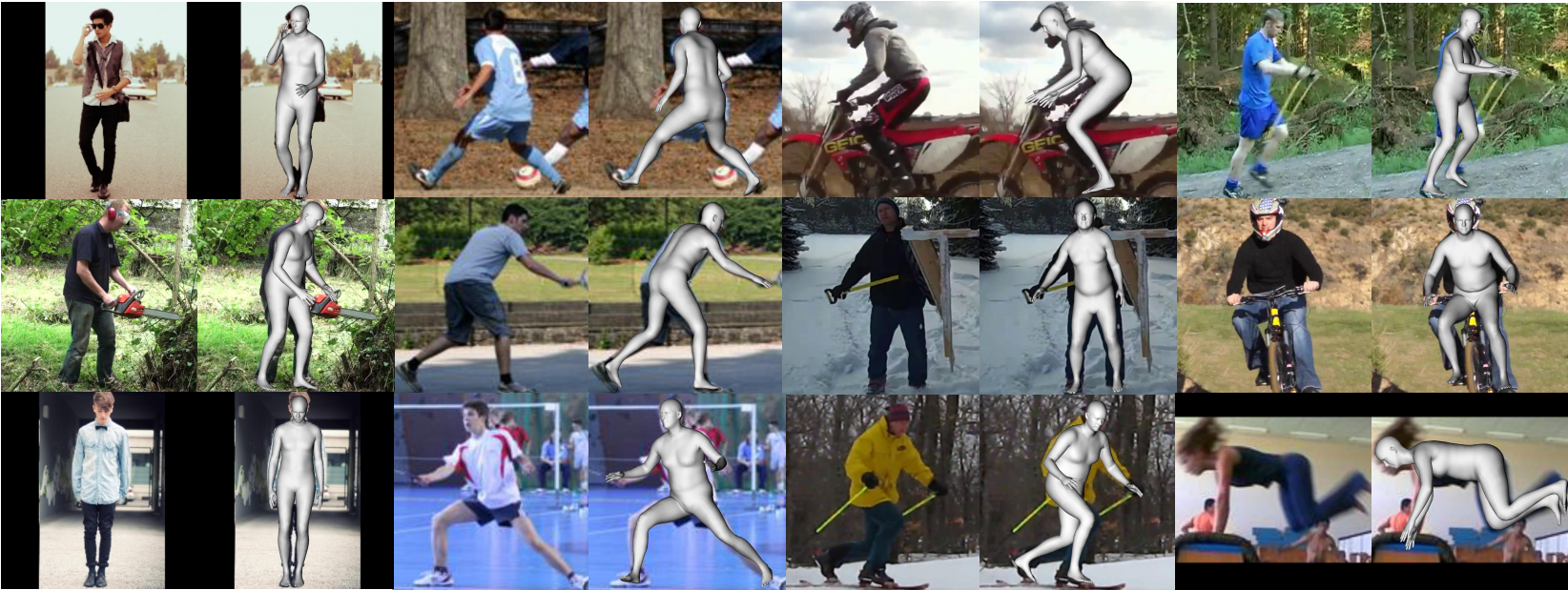}，
\caption{Example results on UP dataset.}
\label{fig:demo}
\end{figure*}

\section{Introduction}
Estimating human pose and shape from a single image is a problem being studied for years, which has many applications in body animation, human action recognition and virtual reality. In the past several years, various approaches have been proposed to either jointly\cite{Varol_2018_ECCV, varol2017learning,kanazawa2018end} or partially\cite{jackson20183d,pavlakos2017coarse,lcrnet} solve this problem. In this work, we present an end-to-end method to directly recover full 3D geometry of human body from a single color image via CNN.

Due to the lack of body shape data, earlier studies focus more on human pose estimation by recovering 3D joint locations from a single image\cite{baseline,ordinal,wild}.  
In the year of 2015, Skinned Multi-Person Linear model (SMPL)\cite{loper2015smpl} was presented, then many methods based on SMPL have been proposed to recover human body shape. In the beginning, researchers used iterative optimization to estimate the pose and shape information from RGB images, typically by fitting a 3D body estimation to some 2D observations, such as 2D keypoints\cite{bogo2016keep} or silhouettes\cite{lassner2017unite}.
These methods usually require a large amount of runtime to solve the optimization problem.
With the emergence of deep learning, many works tried to utilize CNNs to solve human pose and shape estimation in an end-to-end manner\cite{kanazawa2018end,kanazawa2018end,pavlakos2018learning}, and some of them have achieved better performance with faster running speed. 
However, directly predicting human mesh through CNN is challenging, since such training task requires large amount of data with 3D supervision. Recent CNN methods usually predict SMPL parameters and rely on various intermediate 2D representations to guide the training process. 
In these methods, the process of mapping from single RGB image to 3D mesh of human body is decomposed into two stages, that first regresses some types of 2D representations such as joints heatmap, mask or 2D segmentation, then predicts the model parameters from these intermediate results. For instance, \cite{omran2018neural, pavlakos2018learning} improved the performance with the help of joints or segmentation outputs. 
The performance of these works is highly influenced by the choice of intermediate representation and the quality of the output provided by the network solving these sub-tasks. 
Recently, \cite{Varol_2018_ECCV,jackson20183d} argued for an alternative representation for body shape estimation, and demonstrated its effectiveness. In these works, volumetric representations are used as the network output.
In another similar task, 3D face reconstruction, recent work tried to use UV position map\cite{feng2018joint} as a representation and achieved state-of-the-art performance. 

Inspired by \cite{Varol_2018_ECCV,jackson20183d,feng2018joint}, we propose to represent human body in UV space, and train an encoder-decoder network to predict human shape from a single color image. 
The usage of this representation enables an efficient framework to directly regress full 3D mesh of human body, without relying on any intermediate representations. 
We show that our method surpasses other top performing methods in pose and shape estimation with even faster running speed.
The surface and joint errors of our method on SURREAL are 31.5\% and 13.0\% lower than previous state-of-the-art results, respectively. 
On Human3.6M dataset, evaluation on joint error shows that our method performs the best among all previous methods when no additional training data is used, and provides comparable result with the state-of-the-art method focusing on 3D joint estimation in the situation that using additional training data is allowed.
On UP-3D dataset, the performance of our method exceeds previous state-of-the-art methods on both surface and joint errors.

In summary, our main contributions are:

(1) To our best knowledge, we are the first to utilize UV map in addressing the problem of 3D human pose and shape estimation. Based on our 3D representation, we propose an end-to-end framework to predict full human mesh from a single color image.

(2) For the first time in this area, we simplify the neural network training process into one stage, without solving any intermediate sub-tasks such as segmentation or 2D pose estimation, etc.

(3)We train an encoder-decoder network that directly maps the input RGB image to our 3D representation, at extremely fast running speed(200fps).

(4) The performance of our method greatly surpasses other state-of-the-art methods on multiple 3D human body datasets.

\begin{figure*}
\centering
\includegraphics[width=17cm]{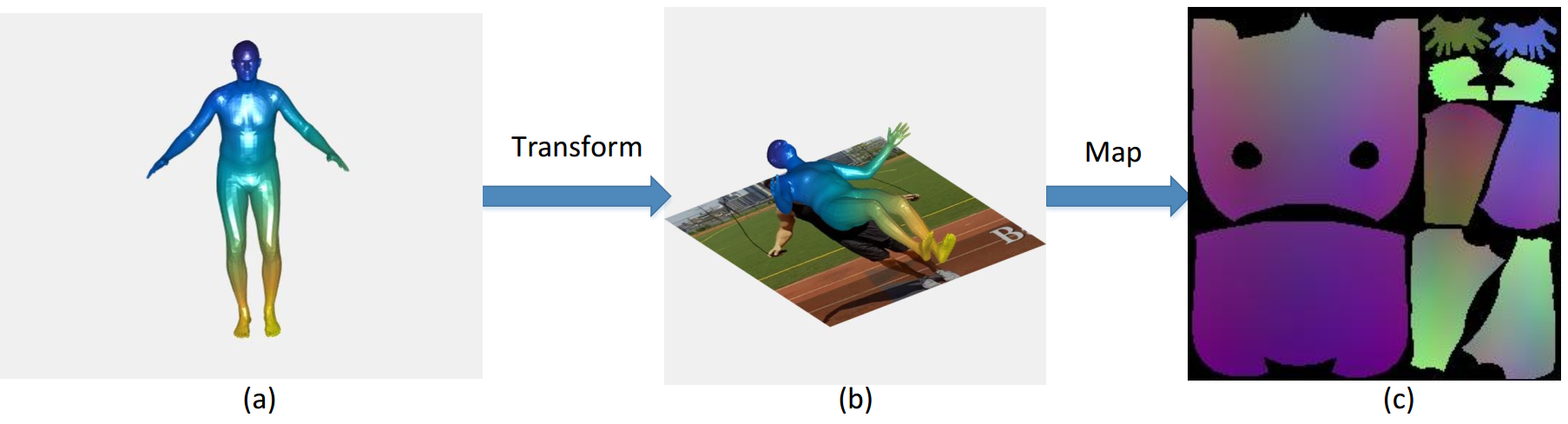}，
\caption{The process of generating UV position map (a): 3D mesh in world coordinate; (b): 3D mesh aligned with the 2D image; (c): generated UV position map.}
\label{fig:uv}
\end{figure*}
\section{Related Works}

\subsection{3D Human pose estimation}
The problem of 3D human pose estimation was commonly simplified as predicting the positions of 3D joints. Methods in this area can be split into two categories with different emphasis on the usage of 2D joints information.

Due to lack of human body datasets with 3D pose label, many papers utilize 2D joint locations to help predict the 3D joint positions. They first predict 2D joint locations using 2D pose detectors\cite{stackedhg, cpm} or use ground-truth 2D pose, then predict the 3D joint positions from the 2D joints information. 
Martinez et al.\cite{baseline} used a simple fully-connected network to regress 3D pose from 2D pose.
Fang et al.\cite{grammar} proposed a bidirectional Recursive Neural Network(RNN) to learn the kinematics grammars between joints and limbs.
As the performance of these two-stage methods highly relies on the accuracy of the 2D joint locations, Stacked Hourglass Network\cite{stackedhg} was used to refine the prediction with a cascaded network architecture.
However, since the final result is regressed from 2D joint locations, the information of the image is not fully utilized.

More recent works directly estimate the 3D joint positions from the input image using CNN.
Due to the depth ambiguity of image, it is very hard to determine whether the limbs are bended forward or backward. To solve the problem, many researchers tried to introduce priors or geometric constraints.
Yang et al.\cite{wild} added an adversarial loss as a prior to avoid unrealistic results. 
In addition to the joints information, Sun et al.\cite{compositional} tried to learn the bone distribution as a geometric constraint. 
Pavlakos et al.\cite{ordinal} adopted the ordinal annotations between joints as complementary information to alleviate the depth ambiguity. 
In this work, we focus on predicting the whole surface geometry of the human body rather than 3D joint positions. 

\subsection{Human shape and pose estimation}
Estimating 3D human pose and shape from a single color image is a challenging problem. Most previous works rely on parametric body models such as SMPL\cite{loper2015smpl} or SCAPE(Shape Completion and Animation for People)\cite{anguelov2005scape} to represent 3D human body and try to predict the parameters of the model.

Earlier works fit the parametric model to some 2D image observations through iterative optimization, which requires very careful initialization, and takes a long time to run.
Bogo et al.\cite{bogo2016keep} firstly proposed a two-stage method: SMPLify, which detects human landmarks through a CNN and then estimates SMPL parameters through iterative optimization. 
Lassner et al.\cite{lassner2017unite} extended SMPLify by minimizing an objective function composed of a data term with 2D keypoints and silhouettes.
Recent methods are usually learning-based and use CNNs to predict the model parameters.
Jun Kai et al.\cite{tan2017indirect} proposed an encoder-decoder architecture to estimate SMPL parameters, with 2D silhouettes as the only supervision.
In Pavlakos's method\cite{pavlakos2018learning}, 2D joint heatmaps and masks were used as intermediate representations to predict SMPL pose and shape parameters respectively. Then a differentiable renderer was used to project 3D mesh to 2D joints and masks, which allows the whole network to be trained end-to-end. 
Similarly, Omran et al.\cite{omran2018neural} used 2D body part segmentation as intermediate representation to estimate SMPL parameters, the effectiveness of other intermediate representations was also explored in this work.

Most recently, \cite{Varol_2018_ECCV} and \cite{jackson20183d} proposed a volumetric representation for the task of 3D human shape estimation. In \cite{Varol_2018_ECCV}, 3D joint positions are predicted as intermediate result and the final body shape is regressed from several intermediate 2D representations.
Differently from all previous methods, we propose to directly predict the 3D human body shape and pose without relying on any parametric body models and intermediate 2D representations.

\subsection{Representations for 3D objects}
For 3D reconstruction tasks, choosing a proper representation of target 3D object is always a critical topic.
When addressing the problem of human pose and shape estimation, the majority of works\cite{omran2018neural, pavlakos2018learning, kanazawa2018end} utilized CNNs to regress the parameters of parametric models like SMPL\cite{loper2015smpl}. 
Recently, \cite{Varol_2018_ECCV,jackson20183d} proposed to use volumetric representation as an alternative for body shape reconstruction. DensePose\cite{alp2018densepose} used a 24 parts UV map to represent the image-to-surface correspondences and regressed body parts coordinates in the UV map for 2D dense pose estimation. 
Regarding generic 3D object reconstruction, representations such as voxel\cite{maturana2015voxnet, yan2016perspective}, octree\cite{tatarchenko2017octree, wang2017cnn, riegler2017octnetfusion} and point cloud\cite{qi2017pointnet++} are commonly used. Similarly, in the case of faces, using model parameters\cite{zhu2016face,Tewari2017MoFA} to represent 3D face is popular. In 3D face reconstruction tasks, volumetric representation\cite{Jackson2017Large} and presenting face surface in UV space\cite{feng2018joint} for reconstructing face geometry have shown some advantages. In this work, we investigate the usage of UV map in human body reconstruction task.

\section{Proposed method}
We identify the representation of human body as a priority for the task of estimating human pose and shape from a single color image, especially when using CNNs to solve this problem. 
Thus, we first introduce our representation of 3D human body for this task in Sec. \ref{repr}.
Then, based on our representation, we propose a framework which uses an encoder-decoder network to directly learn the mapping from a single image to body shape. 
Sec. \ref{network} describes the details of the network and the corresponding losses.
We further explain the implementation details in Sec \ref{details}. 
Finally, we describe the fitting process in Sec \ref{fit smpl}.

\subsection{3D body representation}\label{repr}

Choosing an optimal 3D representation of human body for neural network to regress is an open problem.
Predicting the parameters of body shape model such as SMPL is commonly used by recent approaches. However, the mapping from input image to the low-dimensional parameters is highly non-linear, and it's difficult to train the network with only parameter loss, thus additional losses\cite{kanazawa2018end,omran2018neural,pavlakos2018learning} are combinely used. 
Volumetric representation for human body\cite{Varol_2018_ECCV} is also proposed for this task and demonstrates its effectiveness. The voxel predictions are regularized by additional re-projection loss and other intermediate multi-task supervisions.
Considering that the surface of human body contains most of the information of body shape, and CNN performs well in tasks related to 2D images, 
we propose to record body surface in UV space as the representation of human body geometry in our method.

In the area of human body, UV map, as a surface-to-image representation, is often used to render textures\cite{bogo2014faust, bogo2017dynamic, neverova2018dense}. In this work, we try to utilize UV map to represent the body surface geometry. 
Most 3D human body datasets provide ground-truth defined by SMPL, which provides a default UV map that divides the human body into 10 separate parts. 
DensePose\cite{alp2018densepose} offered an alternative choice of UV map which separates the human body into 24 parts.
We experimented with both candidates and the usage of the default UV map provided better results. Thus we stores the 3D position information of full human body using this UV space.

Specifically,
we first transform the SMPL body model from world coordinate system to camera coordinate system, then spatially align the 3D model with the corresponding color image using orthographic projection, so that the 3D human body model matches the 2D body when projected to the image plane.
For each 3D vertex on the human body surface, the $x$ and $y$ coordinates correspond to the point in the image, the $z$ coordinate is the relative depth to the root point(i.e., the hip point). 
We store the x,y,z coordinates of the vertices as r,g,b color values in the UV map. 
The process of generating the UV position map is shown in Figure \ref{fig:uv}.

\begin{figure}
\centering
\includegraphics[width=3.2in]{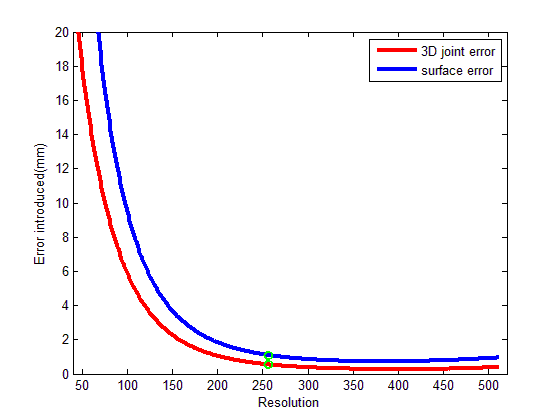}，
\caption{The error introduced by rendering vertices to UV position map and re-sampling vertices from UV position map.}
\label{fig:res_err}
\end{figure}

Notice that the SMPL vertices can be re-sampled from our UV position map, the resolution of the UV position map needs to be properly determined in order to reduce the error. We randomly select several samples of 3D human body to measure the re-sampling errors under different UV map resolutions. The relation between the resolution and the mean re-sampling error is shown in Figure \ref{fig:res_err}. 
Considering that surface error and 3D joint error in state-of-the-art methods are tens of millimeters, we choose 256 as the resolution. The introduced surface or 3D joint error is about 1-2mm which is negligible.

\begin{figure*}
\centering
\includegraphics[width=6.5in]{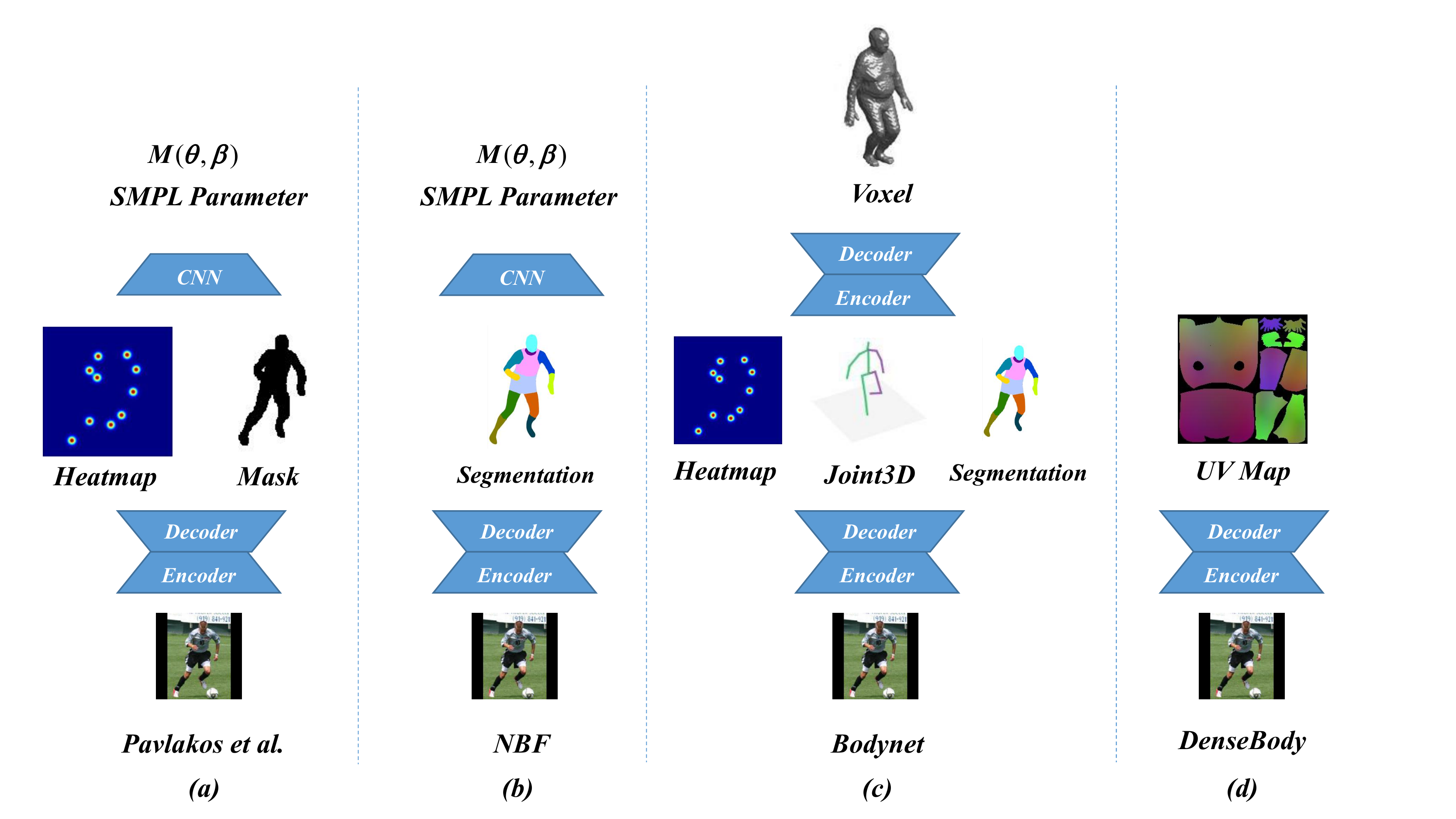}，
\caption{A comparison of several pipelines with the proposed DenseBody pipeline. (a) refers to \cite{pavlakos2018learning}. (b) refers to \cite{omran2018neural}, (c) refers to \cite{Varol_2018_ECCV}. (d) refers to our method.}
\label{fig:pipeline}
\end{figure*}
\subsection{Network and Loss}\label{network}
Our network adopts the encoder-decoder architecture that maps $256\times 256\times 3$ RGB image into $256\times 256\times 3$ UV position map. 
For the encoder part, we use ResNet-18\cite{resnet} considering the balance between the performance and speed. The decoder part is simply composed of 4 consequent up-sampling and convolutional layers. In the ablation study(see Sec. \ref{ablation_loss}), more encoder architectures are evaluated for comparison. 

Unlike previous methods elaborating multiple losses(see Table. \ref{tab:loss}), our loss function computes a straightforward $L_1$ loss between the predicted UV map and the ground-truth. 

\newcommand{\tabincell}[2]{\begin{tabular}{@{}#1@{}}#2\end{tabular}}  
\begin{table}[]
    \centering
    \begin{tabular}{lc}
    \hline
      Method & Loss \\ \hline
      HMR \cite{kanazawa2018end} & \tabincell{c}{$L_{smpl}+L_{joint3d}+L_{joint2d}+$\\$L_{adversarial}$} \\
      NBF \cite{omran2018neural} & \tabincell{c}{$L_{smpl}+L_{joint3d}+L_{joint2d}+$\\$L_{segmentation}$} \\
      Bodynet \cite{Varol_2018_ECCV} & \tabincell{c}{$L_{voxel}+L_{joint3d}+L_{joint2d}+$\\$L_{segmentation}+L_{silhoutte}$} \\ \hline
      DenseBody & $L_{uv\_map}$ \\
    \hline
    \end{tabular}
    \caption{Different losses adopted by other methods.}
    \label{tab:loss}
\end{table}
In order to balance the supervision applied to different body part on the UV position map, we adjust the weight of each point according to the area of the part the point belongs to, so that the network would not over-fit to the body parts with larger area. We employ a weight mask $\beta_{i,j}$ to our loss function, in which the weight is inversely proportional to the part area, thus each part contributes equally to the training process.
In addition, losses of the points around the joints are also emphasized. 

The equation of the weighted $L_1$ loss is shown below.
\[L_{1} =\sum_{j=1}^H{\sum_{i=1}^W{\beta_{i,j} \left(\left| P_{i,j} - P_{i, j}^{gt} \right|\right)}}\]
To encourage spatial smoothness of the output UV map, we follow prior work on image synthesis\cite{tv} and add a total variation regularizer $L_{tv}$:
\[L_{tv} =\sum_k{\sum_{(i,j)\in R_k}{\alpha_k \left(\left| P_{i+1,j} - P_{i, j} \right| + \left| P_{i, j+1} - P_{i, j}\right|\right)}}\]
where $\alpha_k$ adjusts smoothing constraints on different parts, 
and $R_k$ is defined as the $k^{th}$ body part.

Our overall objective is as below, where $\lambda$ is set through validation:
\[L=L_1+\lambda L_{tv}\]

Benefit from our straightforward encoder-decoder network and loss function, our pipeline for human pose and shape estimation is very concise compared to other methods. The comparison of different pipelines is shown in Figure \ref{fig:pipeline}.

\subsection{Implementation details}\label{details}
In pre-processing the training data, all images are cropped and scaled to $256 \times 256$ so that the human body is at the center of the input image with a proper margin between the tight bounding box and the image border. 
We then process the mesh data from 3D body datasets to our proposed UV position map as described in Section \ref{repr}. 
Augmentation is also conducted to improve the performance. The images are randomly translated, rotated, flipped and color-jittered. It should be noted that most augmentations are non-trivial because the corresponding ground-truth should also be transformed.
We implement our method using PyTorch\cite{pytorch}. Adam optimizer\cite{adam} is used for the training process. We set the initializing learning rate as 1e-4 and use 64 as the training batch size. Our training process converges after around 20 epochs. Training on a single GTX 1080 Ti GPU takes around 20 hours. 

\subsection{Fitting the SMPL model}\label{fit smpl}
While our method produces model-free results, for some applications, it is important to produce a parametric model for further manipulation such as retargeting\cite{unseen_pose} or beautification. Moreover, to make a convincing benchmark(see Sec. \ref{performance}), we need to calculate metrics under different joint definitions. However, the 24(or 16) joints commonly adopted in this field can only be calculated through the SMPL model, which necessitates the fitting of SMPL model.

In SMPL model, each body shape can be represented by $V(\theta, \beta) \in R^{3 \times N}$ with $N = 6980$, where $\theta \in R^{3 \times K}, K = 23$ and $\beta \in R^{10}$ denote the parameters for the human pose and shape respectively.  
Our network can provide ordered vertices that each point corresponds to a vertex in SMPL model. 
As mentioned in Section \ref{repr}, there is a similarity transformation from body surface generated in SMPL to surface recorded in UV map, we denote rotation matrix as $R \in R^{3 \times 3}$, translation as $ t \in R^3$ and scale as $ s \in R$, 
the transformed vertices can be represented by
\begin{equation}
\hat{V}(\theta, \beta) = sRV(\theta, \beta) + t
\end{equation}

We define the re-sampled vertices from predicted UV position map as  $\widetilde{V} \in R^{3 \times N}$, then the objective is to find the parameter $\{\widetilde{\theta}, \widetilde{\beta}\}$ that minimizes the weighted L2 between $\hat{V}(\theta, \beta)$ and $\widetilde{V}$:
\[  \{\widetilde{\theta}, \widetilde{\beta}\} = \mathop{\arg\min}_{\theta, \beta} \ \ \sum_{k=1}^N  \|\hat{V_k}(\theta, \beta) -\widetilde{V_k}\|^2_2\]

\begin{figure*}
\centering
\includegraphics[width=17cm]{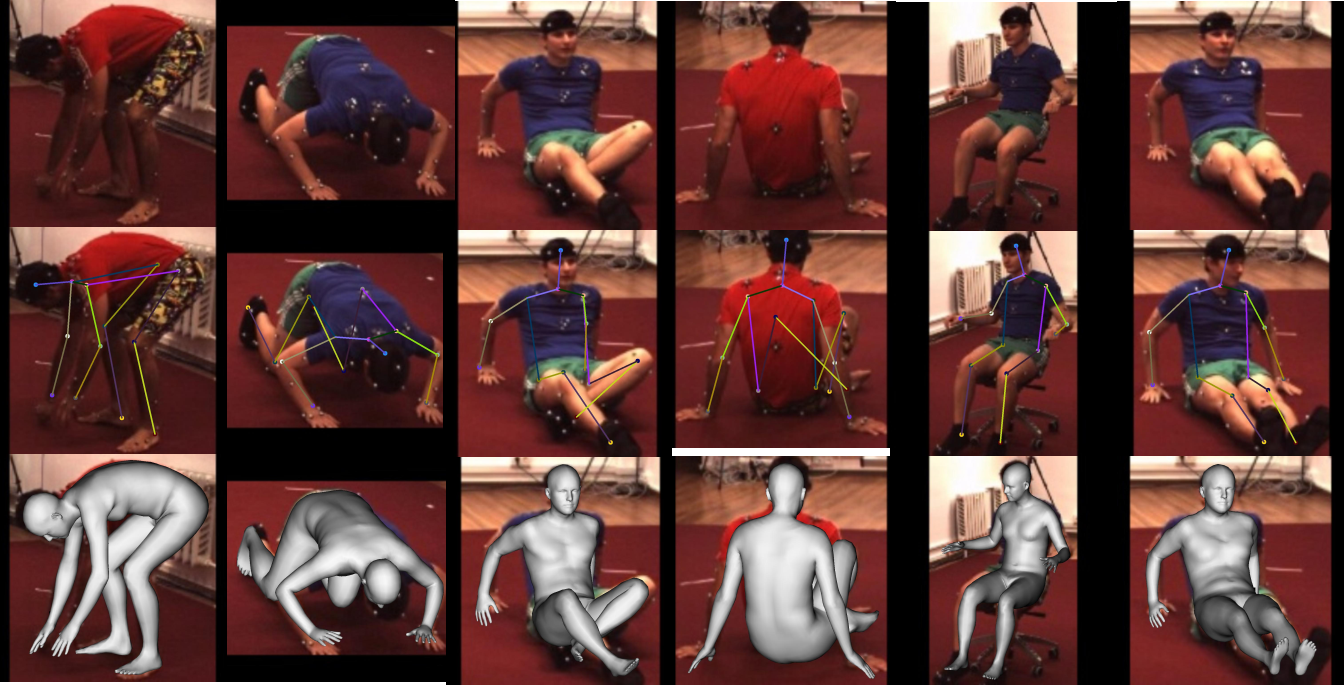}，
\caption{Pose and shape estimation results on Human3.6M.}
\label{fig:h36_demo}
\end{figure*}
Note that the fitting stage should not be considered as some kind of essential post-processing for shape estimation since the reconstruction errors before and after fitting are roughly the same. In the following experiments, the usage of the fitting results depends on the benchmark standards.

\section{Experiments}
In this section we present the evaluation of our method and compare it with previous methods. 
First we describe the datasets used in this work. Next, we provide the evaluation results on these datasets as well as the comparison with other methods in Section \ref{performance}. Moreover, the running speed is also compared in Section \ref{runtime}.
 In the end, we conduct several ablation studies on different issues concerning our method.

\subsection{Datasets}\label{data}
\textbf{SURREAL}. SURREAL dataset\cite{varol2017learning} is a large-scale synthetic dataset with SMPL  parameters\cite{varol2017learning} released in 2017. The original training set consists of 1,964 video sequences of 115 subjects, in a total of 5,342,090 frames. In our experiment, we select one out of three frames for reducing redundancy. Additionally, frames with incomplete human body are discarded. Our final training set includes around 1.6 million frames. 
The test set contains 703 sequences of 30 subjects, including 12528 clips and more than 1 million frames in total. We choose the same sub-sets as in \cite{Varol_2018_ECCV} to test our model for comparison.
Specifically, we extract a sub-testset of 493 frames, which samples part of the test sequences (493 out of 703) and extracts the middle frame of the middle clip of each sequence. We use notation $S1$ for this sub-testset. 
Further more, we notice that some frames in $S1$ contain incomplete bodies, thus we extract the middle frames from all the clips(12528) of the testset and remove those that human body exceeds the image, resulting in 11370 test images.
We denote this sub-testset as $S2$.

\begin{table}[]
    \centering
    \begin{tabular}{clcc}
    \hline
      Testset & Method & Surface Error & Joint Error \\ 
      \hline
      \multirow{4}{*}{\rotatebox{90}{S1}} 
      & SMPLify++\cite{bogo2016keep} & 75.3 & - \\ 
      & Tung et al. \cite{tung} & 74.5 & - \\
      & Bodynet\cite{Varol_2018_ECCV} & 73.6 & 46.1 \\
      & Ours & \textbf{54.2} & \textbf{40.1} \\ 
      \hline
      \rotatebox{90}{S2} & Ours & \textbf{43.3} & \textbf{34.6} \\
    \hline
    \end{tabular}
    \caption{Performance on SURREAL, surface errors and 3D joint errors in mm. } 
    \label{tab:surreal}
\end{table}

\textbf{Human3.6M}. Human3.6M dataset\cite{h36m} contains 15 action sequences of several individuals, captured in a controlled environment. There are totally 1.5 million training images with 3D annotations.
\cite{kanazawa2018end} applied  MoSH\cite{mosh} to the raw 3D MoCap marker data to obtain the ground-truth SMPL parameters. 
Following previous works \cite{lcr,volumetric}, our network is trained on 5 subjects (S1, S5, S6, S7, S8) and tested on 2 subjects(S9, S11). For training, we extract one frame from every five adjacent frames so that there are 300K non-repeating training images. 

\textbf{UP-3D}. UP-3D dataset\cite{lassner2017unite} is obtained by fitting SMPL model on multiple human pose datasets and then cleaned by annotators. It contains 5703 training images, 1423 validation images and 1389 test images.
We report 3D joint error and surface error on the test set and validation set. We also report results on images from LSP subset(T1)\cite{Varol_2018_ECCV} and images with the rotation angle less than $72^{\circ}$ (T2)\cite{Varol_2018_ECCV, tan2017indirect}.

\subsection{Performance}\label{performance}

\begin{table}[]
    \centering
    \begin{tabular}{lcrr}
    \hline
     Method & MPJPE& MPJPE-PA \\ \hline
      $*$Tome et al. \cite{tome} & - & 70.7 \\
      $*$Pavlakos et al. \cite{pavlakos2017coarse} & -&51.9 \\
      $*$Mehta et al. \cite{mehta} & -&54.6 \\
      $*$Sun et al.$^{\dagger}$ \cite{compositional} & 59.1 & 48.3 \\
      $*$Kinauer et al. \cite{kinauer} & - & 50.2 \\
      $*$Martinez et al. \cite{baseline} & 62.9 & 47.7 \\
      $*$Yang et al.$^{\dagger}$ \cite{wild} & 58.6 & \textbf{37.7} \\
      \hline
      SMPLify \cite{bogo2016keep} &-&82.3 \\
      SMPLify++ \cite{lassner2017unite} &-&80.7 \\
      Pavlakos et al. \cite{pavlakos2018learning}&-&75.9 \\
      HMR \cite{kanazawa2018end}   &-&77.6 \\
      HMR$^{\dagger}$\cite{kanazawa2018end} &88.0&56.8 \\
      NBF \cite{omran2018neural} &-&59.9\\
      Bodynet$^{\dagger}$ \cite{Varol_2018_ECCV} &49.0&-\\
      Ours &\textbf{51.0}&\textbf{41.3}\\
      Ours$^{\dagger}$ &\textbf{47.3}&\textbf{38.1} \\ 
      \hline
    \end{tabular}
    \caption{Performance on Human3.6M, MPJPE and MPJPE-PA errors in mm. $*$ indicates methods that only output 3D joints. $\dagger$ indicates methods using extra training data.} 
    \label{tab:h36m}
\end{table}
On SURREAL, we compare both surface error and joint error between our method and previous works with reported performance on this dataset.
In $S1$ testset, three other methods\cite{bogo2016keep,tung,Varol_2018_ECCV} are compared. In particular, we use the errors reported in BodyNet\cite{Varol_2018_ECCV}. 
As shown in Table \ref{tab:surreal}, the surface error of our method is significantly lower than previous state-of-the-art result, with 31.5\% relative improvement. Regarding the 3D joint error, our method outperforms BodyNet\cite{Varol_2018_ECCV} by 13.0\%. 
We also report the performance of our method on $S2$. We believe that $S2$ is more appropriate for evaluating the performance of the methods in this field, due to the reason described in \ref{data}. We can see that both errors greatly decreased when we exclude those incomplete image frames.

\begin{figure*}
\centering
\includegraphics[height=4.7cm,width=17cm]{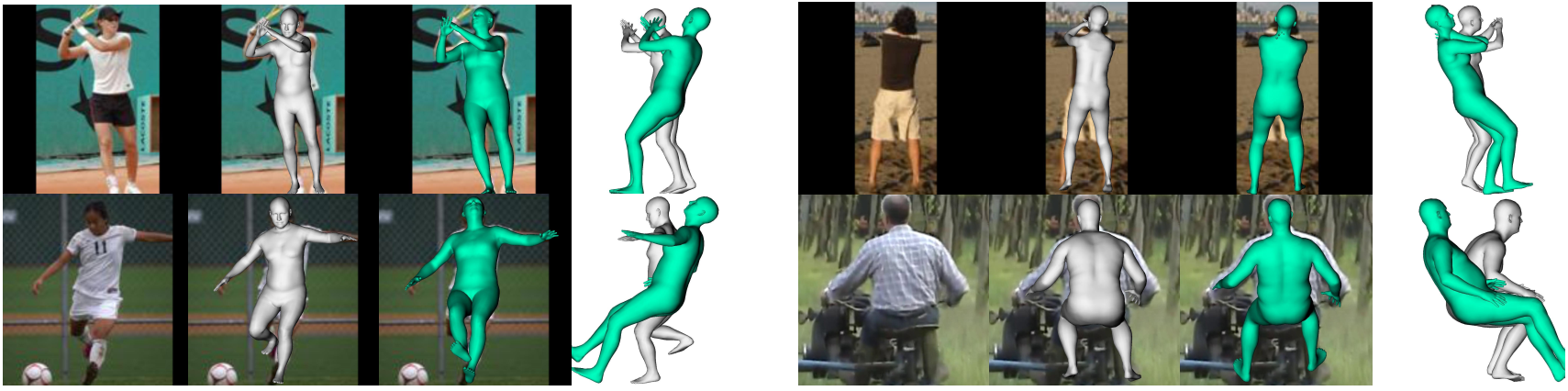}，
\caption{3D body reconstruction results on UP-3D dataset. Gray: our prediction; Green: ground truth. In these cases, the depth information of ground truth is not processed well.}
\label{fig:up_err}
\end{figure*}
We evaluate our method on Human3.6M dataset and conduct comparison to other methods on 3D human body pose estimation, notice that some methods only predict 3D joint positions.
MPJPE(Mean per joint position error) and MPJPE-PA(the MPJPE after Procrustes Alignment) are used as the metrics to measure 3D joint error.
MPJPE refers to the 3D joint error when the hip depth of the  predicted joints are aligned to the hip depth of the ground-truth, and MPJPE-PA refers to the 3D joint error after making a rigid alignment between the prediction and the ground-truth. 
To match the joint definition adopted by previous works, we use the method mentioned in Section \ref{fit smpl} to obtain SMPL parameters, which provide us 16 joints for comparison.

The results are shown in table \ref{tab:h36m}. 
In general, methods focusing on predicting 3D joint positions perform better than methods addressing shape estimation on evaluation of joint errors.
When no extra training data is used, our method surpasses all previous methods whether or not focusing on 3D joint estimation.
In the condition that extra training data could be used, our method also performs better than all other methods dealing with human body shape and pose estimation. 
Even compared with the methods specially focusing on predicting 3D joint positions, our method still produces comparable result. Figure \ref{fig:h36_demo} shows qualitative results on Human3.6M.

\begin{table}[]
    \centering
    \begin{tabular}{clcc}
    \hline
      Testset & Method & Surface Error & Joint Error \\ 
      \hline
      \multirow{2}{*}{\rotatebox{90}{T1}} & Bodynet\cite{Varol_2018_ECCV} & 102.5 & 83.3 \\ 
       & Ours & \textbf{96.4} & \textbf{76.3} \\
      \hline
       \multirow{3}{*}{\rotatebox{90}{T2}}& Direct learning\cite{tan2017indirect} & -- & 105.0 \\
       & Bodynet\cite{Varol_2018_ECCV} & 80.1 & 69.6 \\
       & Ours & \textbf{79.8} & \textbf{65.8} \\
      \hline
      \multirow{2}{*}{\rotatebox{90}{Val}} & NBF\cite{omran2018neural} & -- & 82.0 \\
       & Ours & \textbf{96.3} & \textbf{77.1}\\
      \hline
      \multirow{3}{*}{\rotatebox{90}{Test}} & Lassner et al.\cite{pavlakos2018learning}& 169.8 & --\\
      & Pavlakos et al.\cite{pavlakos2018learning}& 117.7 & --\\
      & Ours & \textbf{91.7} & \textbf{71.4}\\
    \hline
    \end{tabular}
    \caption{Performance on UP-3D, surface errors and 3D joint errors in mm.} 
    \label{tab:up-3d}
\end{table}

The results on UP-3D are shown in Table \ref{tab:up-3d}. Obviously, the performance of our method is better than other state-of-the-art methods on all testsets.
We also notice that the error on the dataset UP-3D is higher than other datasets, thus we investigate some test images which have large test errors(about 200-300 mm). The visualization examples are presented in Figure \ref{fig:up_err}. We can see that, our predictions are more accurate than the ground truth in some cases.

\subsection{Running time}\label{runtime}
Inference runtime is also an important metric, and we make a comparison between our method and other four state-of-the-art works, HMR \cite{kanazawa2018end}, NBF \cite{omran2018neural}, Bodynet \cite{Varol_2018_ECCV} and \cite{pavlakos2018learning}. 
For method in \cite{pavlakos2018learning}, we use the running time result reported in the paper, which is evaluated on TITAN X. For other methods, we use the official implementation publicly released by the authors and evaluate all the methods on GTX 1080 Ti. The inference runtime performance is compared in Table \ref{tab:runtime}. Apparently, our method is significantly faster than other methods.
\begin{table}[]
    \centering
    \begin{tabular}{lr}
    \hline
     Method    & Running time \\ \hline
      HMR \cite{kanazawa2018end} & 1270 \\
      NBF \cite{omran2018neural} & 169 \\
      Bodynet \cite{Varol_2018_ECCV} & 1810  \\
      Pavlakos$^1$ \cite{pavlakos2018learning} & 50  \\
      Ours & \textbf{5} \\
      \hline
    \end{tabular}
    \caption{Single forward runtime on the GTX 1080 Ti, in millisecond. $^1$ is conducted on TITAN X.} 
    \label{tab:runtime}
\end{table}

\subsection{Ablation Study}\label{ablation}
In this section, we conduct several ablation studies to evaluate the influence of different components in our framework.
To investigate whether network architecture is an importance factor for the performance, we trained our network with different encoder network architectures, and compare their performance on Human3.6M.
We also examine the options of various loss functions which are commonly used in this field. 

\subsubsection{Network Architecture}\label{ablation_network}
In our method we use ResNet-18 for the encoder module, we also experiment with VGG11\cite{vgg}, ResNet-34\cite{resnet} and MobileNet\cite{mobilenet} for comparision.
The results on Human3.6M are shown in Table \ref{tab:encoder}. 
To measure the computational cost of each architecture, the Giga-floating-point operations (GFLOPs) in the number of multiply-adds are also appended. 
It is obvious that no matter what encoder we use, our model performs similarly and always outperforms previous methods. It should be noted that all the models are trained from the scratch without special tuning process. 
The results prove that network architecture of encoder is not a crucial issue for the performance of our proposed method.

\begin{table}[]
    \centering
    \begin{tabular}{lrr}
    \hline
     Encoder Architecture    & MPJPE & GFLOPs \\ \hline
      VGG11 \cite{vgg} & 50.6 & 16 \\
      ResNet-34 \cite{resnet} & 49.6 & 4 \\
      ResNet-18 \cite{resnet} & 51.0 & 2 \\
      MobileNet \cite{mobilenet} & 52.8 & 0.57 \\
      \hline
    \end{tabular}
    \caption{Ablation study on network architecture. 3D joint errors in mm.}
    \label{tab:encoder}
\end{table}

\subsubsection{Loss}\label{ablation_loss}
To investigate the performance of different loss functions, we trained our network using different loss terms, including L1 loss on the UV position map, 3D joint loss, adversarial loss, and different weight masks applied to the L1 loss function.
The performance is evaluated on Human3.6M using MPJPE and the results are shown in Table \ref{tab:ablation loss}. We can see that adding more loss terms to the loss function does not introduce significant improvement to the result. Besides, using weight mask shows better effect than adding extra loss term.
We suppose our 3D representation has the advantage of incorporating various labeled components and makes it convenient to apply weights to different loss terms.

\begin{table}[]
    \centering
    \begin{tabular}{lr}
    \hline
     Loss    & MPJPE \\ \hline
     $L_1$ & 65.0 \\
     $L_1$ + $L_{adv}$ & 60.3 \\
     $L_1$ + $L_{joint}$ & 57.3 \\
     $L_1$ w/ $w_{part}$ & 56.1 \\
     $L_1$ w/ $w_{part}$ w/ $w_{joint}$ & 51.4 \\
     $L_1$ w/ $w_{part}$ w/ $w_{joint}$ + $L_{tv}$ & 51.0 \\
      \hline
    \end{tabular}
    \caption{Ablation study on different loss functions, 3D joint errors in mm.}
    \label{tab:ablation loss}
\end{table}

\section{Conclusion}
In this paper, we propose DenseBody, a novel method for 3D human pose and shape estimation from a single color image. 
We first develop our 3D representation in the UV space, and use a UV position map to represent the 3D human mesh. Then we train an encoder-decoder network to directly regress the map from input RGB image.
We also demonstrate that without using any intermediate representations, the task of 3D human body pose and shape estimation can be directly solved with superior performance on several 3D human body datasets. 
We believe DenseBody provides a simple and efficient framework for future works related to 3D human body, given its performance and extremely fast running speed. 


{\small
\bibliographystyle{ieee}
\bibliography{densebody}
}

\end{document}